\documentclass[10pt,twocolumn,letterpaper]{article}

\usepackage[pagenumbers]{cvpr}

\usepackage[utf8]{inputenc}

\usepackage{graphicx}
\usepackage{amsmath, amsfonts}
\usepackage{booktabs}
\usepackage{multirow, array}
\usepackage{mathtools}
\usepackage{xcolor}
\usepackage{url}
\usepackage{microtype}

\usepackage{enumitem}
\usepackage{pifont}
\usepackage{dsfont}
\usepackage{bbm}
\usepackage{bm}
\usepackage{arydshln}
\usepackage{makecell}
\usepackage{xspace}

\usepackage{tikz}
\usepackage[most]{tcolorbox}

\usepackage{algorithm}
\usepackage{algpseudocode}
\algnewcommand{\LineComment}[1]{\State \# #1}

\newcommand{\cmark}{\ding{51}}

\usepackage{booktabs}
\usepackage[table]{xcolor}
\usepackage{makecell}
\definecolor{colhl}{RGB}{255,255,255}
\definecolor{rowhl}{gray}{0.95}

\usepackage[normalem]{ulem}
\newcommand{\best}[1]{\textbf{#1}}
\newcommand{\second}[1]{\uline{#1}}

\usepackage{amsmath}
\usepackage{listings}
\usepackage{booktabs,arydshln}
\usepackage{graphicx}
\usepackage{adjustbox}

\AtBeginEnvironment{table}{\normalsize}
\AtBeginEnvironment{table*}{\normalsize}

\usepackage{tcolorbox}
\tcbuselibrary{breakable}

\definecolor{cvprblue}{rgb}{0.21,0.49,0.74}
\usepackage[
    pagebackref,
    breaklinks,
    colorlinks,
    allcolors=cvprblue
]{hyperref}

\def\paperID{35160}
\def\confName{CVPR}
\def\confYear{2026}

\title{Generate, Analyze, and Refine: Training-Free Sound Source Localization via MLLM Meta-Reasoning}

\author{
Subin Park \quad Jung Uk Kim\thanks{Corresponding author}\\
Kyung Hee University\\
{\tt\small \{subin.park, ju.kim\}@khu.ac.kr}
}

\begin{document}
\maketitle

\begin{abstract}
Sound source localization task aims to identify the locations of sound-emitting objects by leveraging correlations between audio and visual modalities. Most existing SSL methods rely on contrastive learning-based feature matching, but lack explicit reasoning and verification, limiting their effectiveness in complex acoustic scenes. Inspired by human meta-cognitive processes, we propose a training-free SSL framework that exploits the intrinsic reasoning capabilities of Multimodal Large Language Models (MLLMs). Our Generation--Analysis--Refinement (GAR) pipeline consists of three stages: Generation produces initial bounding boxes and audio classifications; Analysis quantifies Audio-Visual Consistency via open-set role tagging and anchor voting; and Refinement applies adaptive gating to prevent unnecessary adjustments. Extensive experiments on single-source and multi-source benchmarks demonstrate competitive performance. The source code is available at \url{https://github.com/VisualAIKHU/GAR-SSL}.
\end{abstract}

\section{Introduction}
\label{sec:main_final}

Sound Source Localization (SSL) task aims to identify the locations of sound-emitting objects within an image by leveraging the correlation between audio and visual information \cite{iterative_contrastive_learning, Mix_and_localize, Learning_to_localize_sound_source, Localizing_visual_sounds_hardway, Hear_the_flow, Discriminative_selfsupervised, NoPrior, Exploiting_transformation, weakly_supervised, Audio_visual_grouping, Multiple_ssl_coarse_fine, Audio_visual_spatial_integration, proposal_based_paradigm}. Its ability to ground sounds in the visual scene makes the SSL task a crucial technology across diverse applications, including autonomous navigation \cite{Sim2real, UAV_Embedded_Microphone}, human-robot interaction \cite{Human_Robot}, and surveillance systems \cite{Drones_and_UAVs}.

Existing SSL research has primarily followed two directions: single-source and multi-source localization approaches. Single-source methods predominantly rely on contrastive learning \cite{Mix_and_localize, negative_aware_contrastive_learning, contrastive_learning_representations}, focusing on improving positive sample quality \cite{Learning_sound_localization_similar_samples}, iterative learning \cite{iterative_contrastive_learning}, and enhanced negative sample handling \cite{negative_samples}. Multi-source approaches have explored pseudo-label learning \cite{iterative_contrastive_learning}, graph-based object relationship modeling \cite{Mix_and_localize}, and iterative audio-visual correspondence discovery. Recently, Um et al. \cite{OA_SSL} adopted Multimodal Large Language Models (MLLMs) as auxiliary components for training vision models (\textit{e.g.,} ResNet18 \cite{resnet18}).

However, the above-mentioned methods share a fundamental limitation: they treat the SSL task solely as a feature matching problem. They primarily focus on aligning audio and visual embeddings without verifying whether the matched region corresponds to the sound source or performing any causal or semantic reasoning. In contrast, humans engage in a multi-step reasoning process \cite{multimodal_chain_of_thought, corvid_mllm_cot_reasoning, reasoningtrack, orderchain} when localizing sound sources. They (\textit{i}) first perceive the characteristics of auditory and visual signals, (\textit{ii}) systematically analyze each candidate object, and (\textit{iii}) then refine their final conclusions. This process goes beyond simple matching, involving meaningful interpretation and verification.

\begin{figure}
    \centering
    \includegraphics[width=1.0\linewidth]{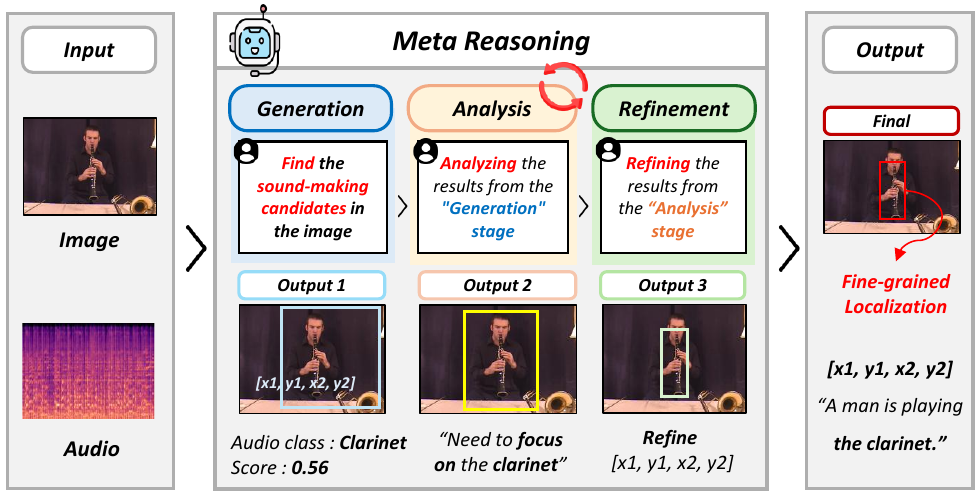}
    \vspace{-0.4cm}
    \caption{
    Overview of the proposed \textbf{Generation-Analysis-Refinement Sound Source Localization (GAR-SSL)} framework. Given an image-audio pair, the model performs three meta-reasoning steps: Generation produces an initial bounding box and audio label, Analysis evaluates Audio-Visual Consistency through role-based reasoning, and Refinement adjusts the localization to obtain a fine-grained final bounding box. This process enables explainable and training-free audio-visual localization.
    }
    \label{fig:figure1}
    \vspace{-0.4cm}
\end{figure}

Recently, Multimodal Large Language Models (MLLMs) have demonstrated strong capabilities in cross-modal understanding, structured reasoning, and instruction following \cite{Meta-reasoning, multimodal_chain_of_thought, llm_visual_reasoning_coord, mm_reason, human_centered_mllm, mllm_survey, survey_on_mllm, corvid_mllm_cot_reasoning, reasoningtrack, orderchain}. These models can interpret complex visual scenes, integrate information across modalities, and execute multi-step reasoning guided by natural-language prompts. Their robust zero-shot generalization and inherent reasoning abilities make them a promising tool for sound source localization.

In this paper, inspired by the human cognitive process of sound source reasoning, we propose a training-free, zero-shot SSL framework that equips MLLMs \cite{mllm_search_zero_shot} with human-like meta-reasoning capabilities \cite{Meta-reasoning}. Rather than treating SSL as a simple feature matching task, we reformulate it as a structured cognitive reasoning procedure composed of three stages-Generation, Analysis, and Refinement-that operate in a coarse-to-fine manner, as shown in Figure \ref{fig:figure1}. Each stage plays a distinct and complementary role. Specifically, Generation broadly enumerates plausible sound-emitting candidates and produces an initial spatial hypothesis; Analysis then verifies each candidate by evaluating Audio-Visual Consistency through role-based reasoning and anchor voting; and Refinement integrates the verification results to correct localization errors and produce a fine-grained final bounding box. Together, these three stages form an explainable and training-free audio-visual localization pipeline, as shown in Figure \ref{fig:figure1}.

\textbf{(\textit{i}) In the Generation stage}, the MLLMs broadly interpret audio characteristics (\textit{e.g.,} pitch, timbre, rhythm) and identifies all visually present objects to enumerate every plausible sound-emitting candidate. Unlike prior approaches that immediately match audio to a single region, this coarse reasoning step keeps the hypothesis space wide to avoid missing potential sources. For example, when a knocking sound is heard, MLLMs consider not only drums but also cymbals, clapping hands, tables, and any other object that could produce a hit-like sound.

\textbf{(\textit{ii}) In the Analysis stage}, the MLLMs then perform fine-grained verification of each candidate using two complementary checks: physical plausibility, which evaluates whether the object can realistically produce the sound, and audio-visual semantic consistency, which examines whether the predicted object is semantically consistent with the input audio signal. This dual verification removes visually salient but irrelevant objects. Unlike simple feature matching, the MLLMs also provide causal explanations and confidence scores, mimicking how humans evaluate plausibility.

\textbf{(\textit{iii}) In the Refinement stage}, the MLLM finally integrate all verification results to compare remaining hypotheses and make a context-aware final decision, considering cues such as volume-distance consistency and scene semantics. It revisits early assumptions and corrects errors when needed, enabling the model to reach a stable and reliable sound source localization outcome.

We summarize our main contributions as follows:
\begin{itemize}
    \item We propose a simple yet effective training-free SSL framework that exploits MLLMs meta-reasoning through a Generation-Analysis-Refinement pipeline.
    \item We introduce an open-set role tagging and anchor voting mechanism that explicitly identifies sound-producing components and quantifies spatial confidence, yielding an interpretable and verifiable reasoning process.
    \item We design an adaptive gating mechanism to decide when refinement truly improves predictions, preventing performance degradation from unnecessary adjustments.
    \item Experimental results on VGGSound and MUSIC datasets demonstrate the effectiveness of the proposed method for both single-source and multi-source localization.
\end{itemize}
 
\section{Related Work}
\label{sec:main_final}

\begin{figure*}
    \centering
    \includegraphics[width=\linewidth]{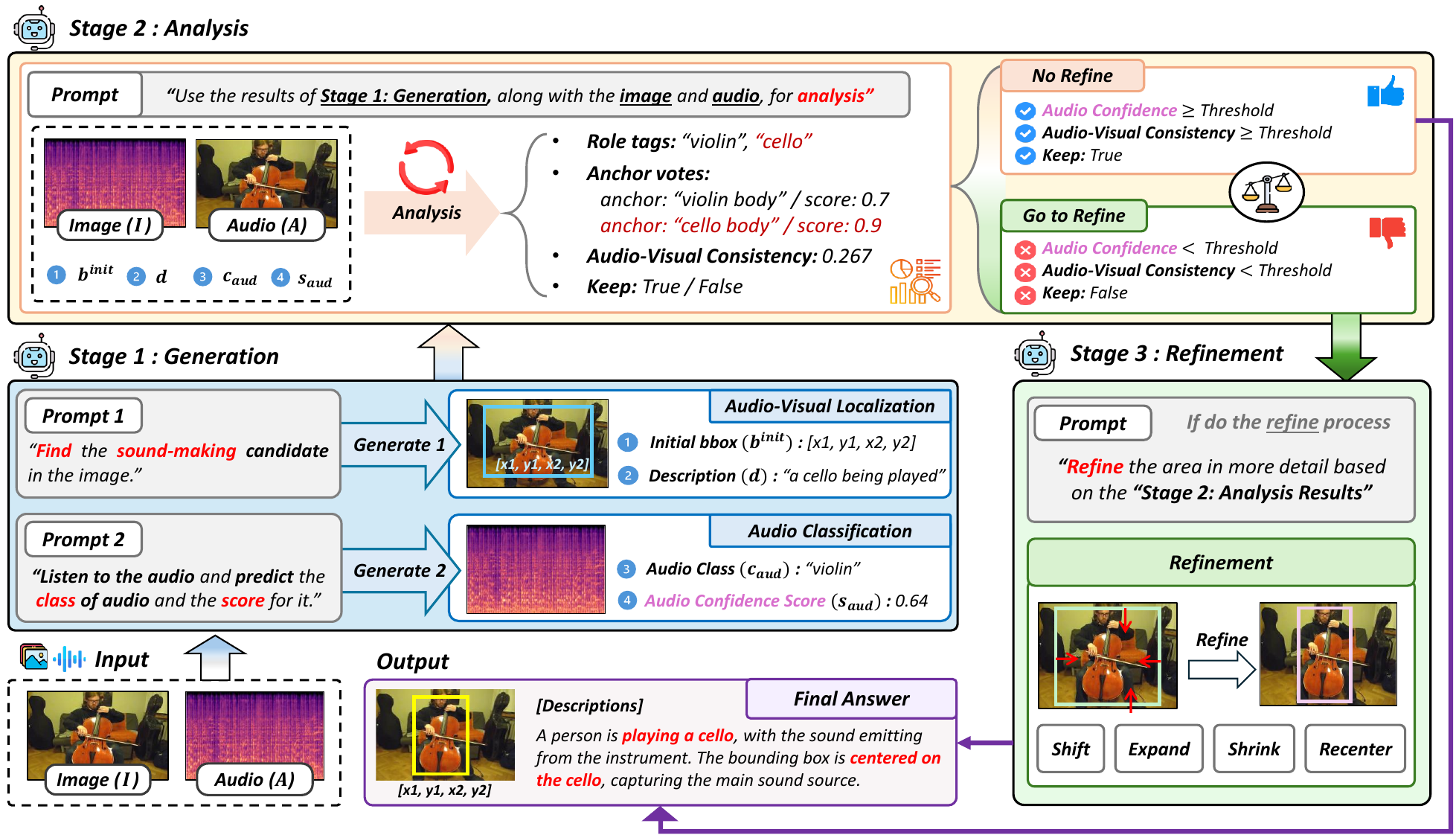}
    \vspace{-0.5cm}
    \caption{The proposed training-free framework consists of three stages: (\textit{i}) Generation produces initial bounding boxes and audio classifications from image-audio pairs; (\textit{ii}) Analysis evaluates consistency through role tagging, anchor voting, and scoring, repeated $N$ times for consensus; (\textit{iii}) Refinement applies adaptive gating and geometric operations to adjust localization. All operations are performed via MLLMs prompt engineering without training.}
    \label{fig:figure2}
    \vspace{-0.2cm}
\end{figure*}

\subsection{Sound Source Localization}
Sound Source Localization (SSL) aims to infer the positions of sound-emitting objects by integrating auditory and visual information. Existing research generally follows two directions: single-source and multi-source localization. 

Single-source approaches have evolved from early attention-based dual-stream models \cite{Audio_visual_scene_analysis, Learning_to_localize_sound_source} to contrastive learning frameworks \cite{Learning_sound_localization_similar_samples, Localizing_visual_sounds_hardway}, with improvements through pseudo-label refinement, optical-flow guidance, and semantic alignment \cite{iterative_contrastive_learning, CoT-PL, Hear_the_flow, OA_SSL, Audio_visual_spatial_integration}. For multi-source scenarios, prior work has explored coarse-to-fine separation, relational modeling, and discriminative supervision \cite{Discriminative_selfsupervised, Multiple_ssl_coarse_fine, Mix_and_localize}. However, most methods rely heavily on similarity-based matching, which struggles with silent objects, off-screen sounds, and complex acoustic scenes \cite{Mix_and_localize, Multiple_ssl_coarse_fine, NoPrior, OA_SSL}. 

Recent text-guided and MLLM-assisted SSL methods \cite{T_vsl, open-vocabulary_localization} attempt to incorporate semantic cues, yet typically use MLLMs only as auxiliary encoders without leveraging their full reasoning capability.

\subsection{Reasoning in MLLMs}
Recent audio-visual learning research has expanded beyond localization to broader multimodal tasks such as audio-visual speech recognition and joint audio-video generation \cite{review_audio_visual_speech_recognition, Mm-diffusion}. Multimodal Large Language Models (MLLMs) integrate information across modalities and enable structured reasoning beyond traditional vision--language systems \cite{mllm_survey, survey_on_mllm}. Techniques such as multimodal Chain-of-Thought (CoT) reasoning and fine-grained spatial-temporal understanding further enhance structured inference across modalities \cite{multimodal_chain_of_thought, llm_visual_reasoning_coord, Llava-st, corvid_mllm_cot_reasoning}. In-context learning \cite{Flamingo, Gpt-4o} further improves their ability to interpret complex scenes.

Despite this, their application to SSL remains limited. Existing attempts largely focus on zero-shot inference or extraction of auxiliary features \cite{zeroshot_fewshot}, and their effectiveness for SSL remains limited compared to supervised task-specific approaches reported in the literature \cite{Learning_to_localize_sound_source, Localizing_visual_sounds_hardway, NoPrior}. This suggests that previous work has not fully utilized the semantic understanding and cross-modal reasoning capabilities of MLLMs. Motivated by this gap, we reinterpret SSL as a cognitive reasoning process. We structure SSL into generation, analysis, and refinement stages, exploiting the intrinsic reasoning ability of MLLMs. This enables effective training-free localization.

\section{Proposed Method}
\label{sec:main_final}
We propose a training-free three-stage self-refinement framework for Audio-Visual Sound Source Localization (AV-SSL). Our method explicitly models the consistency between visual and audio modalities and performs progressive refinement accordingly. The framework is shown in Figure \ref{fig:figure2}. All stages are implemented through prompt engineering without additional training, generating structured JSON outputs. This enables training-free SSL by directly leveraging the intrinsic cross-modal reasoning and semantic knowledge of MLLMs. Details are in the following subsections.
\subsection{Stage 1: Generation}
The Generation stage aims to produce initial predictions from both visual and audio modalities. Given an image-audio pair $(I, A)$ from the same scene, this stage generates two complementary outputs: (\textit{i}) Audio--Visual Localization, which yields an initial bounding box and a short visual description, and (\textit{ii}) Audio Classification, which predicts an open-vocabulary audio label with an internally estimated confidence score. These two outputs are generated independently and their consistency is assessed in the Analysis Stage to enable more accurate localization. \\

\noindent{\textbf{Audio-Visual Localization.}}
This component performs cross-modal grounding to identify the primary sound source in the visual scene. Given an image-audio pair $(I, A)$, where $I$ denotes the input image and $A$ denotes the input audio, the model first predicts a bounding box:
\\[-0.5em]
\begin{equation}
    \begin{gathered}
    \mathrm{\textit{b}}^{\text{init}} = [x_1, y_1, x_2, y_2],\\
    0 \le x_1 < x_2 \le W,\quad 0 \le y_1 < y_2 \le H,
    \end{gathered}
\label{eq:bbox}
\end{equation}
where $(x_1, y_1)$ and $(x_2, y_2)$ are the top-left and bottom-right coordinates of the bounding box, respectively, and $W$ and $H$ represent the width and height of the image. In addition, model generates a concise natural-language description $d$ of the predicted bounding box to facilitate clearer understanding in the Analysis stage. We denote the localization mapping as:
\begin{equation}
  f_{\mathrm{loc}}(I, A) = ({\mathrm{\textit{b}}^{\text{init}}}, d),
  \label{eq:loc_mapping}
\end{equation}
where $f_{\mathrm{loc}}$ in \cref{eq:loc_mapping} represents the localization function that maps the image-audio pair to the bounding box (\cref{eq:bbox}) and description. 
The key mechanism is cross-modal grounding: audio events are semantically aligned with visually plausible emitters to produce $\mathrm{\textit{b}}^{\text{init}}$, providing a spatial hypothesis for subsequent refinement. \\

\noindent{\textbf{Audio Classification.}} This component provides semantic constraints for localization by analyzing the audio signal independently. Given the input audio $A$, the model predicts an open-vocabulary audio label and a confidence score:
\begin{equation}
    \begin{gathered}
    f_{\mathrm{aud}}(A) = (c_{\mathrm{aud}},\, s_{\mathrm{aud}}),\\
    c_{\mathrm{aud}} \in \mathcal{C}_{\mathrm{open}},\; s_{\mathrm{aud}} \in [0,1],
    \end{gathered}
    \label{eq:audio_class}
\end{equation}
where $f_{\mathrm{aud}}$ is the audio classification function, $c_{\mathrm{aud}}$ is the predicted audio class label, $s_{\mathrm{aud}}$ is the confidence score, and $\mathcal{C}_{\mathrm{open}}$ is an unbounded label space (\textit{e.g.,} free-form strings such as \textit{``violin''}, \textit{``dog barking''}, \textit{``drum roll''}). The scalar $s_{\mathrm{aud}}$ in \cref{eq:audio_class} quantifies the certainty self-reported by the model; higher values indicate clearer acoustic evidence. This classification provides class-level priors about the sound source that complement the spatial localization.
Collecting the above, Generation stage returns:
\begin{equation}
  \mathrm{\textit{Gen}}_\text{out} \;=\; \bigl({\mathrm{\textit{b}}^{\text{init}}},\; d,\; c_{\mathrm{aud}},\; s_{\mathrm{aud}}),
  \label{eq:stage1_output}
\end{equation}
where $\mathrm{\textit{Gen}}_\text{out}$ in \cref{eq:stage1_output} denotes the output of Stage 1 (Generation), consisting of the bounding box $\mathrm{\textit{b}}^{\text{init}}$, visual description $d$, audio class label $c_{\mathrm{aud}}$, and confidence score $s_{\mathrm{aud}}$. These outputs serve as the foundation for Stage 2 (Analysis). In particular, the visual description $d$ and the audio class label $c_{\mathrm{aud}}$ provide complementary semantic cues about the likely sound source, which help the model reason beyond visual silency alone. Meanwhile, $s_{\mathrm{aud}}$ participates in the gating rule (\cref{eq:gating}) of Stage 3 (Refinement).
\subsection{Stage 2: Analysis}

The Analysis stage serves as a reasoning bridge between the initial prediction and the final refinement. Its purpose is to evaluate the consistency between the outputs of the Generation stage and provide detailed guidance for refinement. Given the Generation stage outputs $\mathrm{\textit{Gen}}_\text{out}$ (\cref{eq:stage1_output}) and the input pair $(I, A)$, this stage produces semantic role tags $\mathcal{T}_{\text{role}}$, anchor evidences $\mathcal{A}_\text{anchor}$, an Audio-Visual Consistency score $S_{\mathrm{av}}$, and a keep flag (\textit{k}). Unlike simple binary judgments, it identifies which parts to adjust, why, and how, providing targeted guidance for the Stage 3 (Refinement). \\

\noindent{\textbf{Open-set Role Tagging.}} It identifies the semantic structure of the sound source by discovering functionally relevant parts. Given the image-audio pair $(I, A)$ and the Stage~1 audio label $c_{\mathrm{aud}} \in \mathcal{C}_{\mathrm{open}}$ from \cref{eq:audio_class}, we define a function $f_{\mathrm{role}}$ that contextually discovers roles (parts) directly related to sound generation. The resulting set is written as:
\begin{equation}
    \begin{gathered}
        \mathcal{T}_{\text{role}} \;=\; f_{\mathrm{role}}(I, A, c_{\mathrm{aud}})
        \;\subseteq\; \mathcal{T}_{\mathrm{open}}, \\
        \qquad |\mathcal{T}_{\text{role}}| \in \{0,1,2,3,4\}, 
    \end{gathered}
    \label{eq:role_tagging}
\end{equation}
where $f_{\mathrm{role}}$ is the role discovery function, $\mathcal{T}_{\text{role}}$ is the set of discovered roles, $\mathcal{T}_{\mathrm{open}}$ denotes an open role vocabulary without predefined categories, and $|\mathcal{T}_{\text{role}}|$ is the cardinality of the set $\mathcal{T}_{\text{role}}$ (the number of roles). In our implementation, the maximum number of roles is set to 4 as a design hyperparameter. To ensure that tags correspond to verifiable visual evidence, we impose a visibility constraint requiring every selected role to be observable in the current frame:
\begin{equation}
\mathrm{vis}(t \mid I)=1
\quad \text{for all } t \in \mathcal{T}_{\text{role}},
\label{eq:visibility}
\end{equation}
where $t$ is an individual role tag belonging to $\mathcal{T}_{\text{role}}$, and $\mathrm{vis}(t\mid I)$ is a function representing the visibility of role $t$ in image $I$, where 1 indicates observability. These role tags, which \cref{eq:role_tagging} satisfy the visibility constraint (\cref{eq:visibility}), provide structural constraints that guide the refinement process toward semantically meaningful sound-making components. \\

\noindent{\textbf{Anchor Voting.}} It identifies visual evidence of sound source to assess localization quality. Given $(I, A, c_{\mathrm{aud}}, \mathrm{\textit{b}}^{\text{init}})$, we define an anchor voting function that produces semantic anchors and their confidence scores based on semantic evidence rather than direct coordinate prediction:
\begin{equation}
    \begin{aligned}
    \mathcal{A}_\text{anchor} \;=\;&\;
    f_{\mathrm{anchor}}(I, A, c_{\mathrm{aud}},\mathrm{\textit{b}}^{\text{init}}),
    \\[2pt]
    =\;&\; \{(a_i,\; s_i)\}_{i=1}^{m},
    \qquad
    m \in \{0,1, 2,\dots,5\},
    \end{aligned}
    \label{eq:anchor_voting}
\end{equation} 
where $\mathcal{A}_\text{anchor}$ is the anchor voting result set, $f_{\mathrm{anchor}}$ is the anchor voting function, and $m$ is the number of discovered anchors. In our implementation, the maximum number of anchors is set to 5 as a design hyperparameter. Each anchor is defined as follows:
\begin{equation}
    a_i \in \mathcal{A}_{\mathrm{open}}, 
    \qquad
    s_i \in [0,1],
    \label{eq:anchor_def}
\end{equation}
where $a_i$ denotes the $i$-th semantic anchor (\textit{e.g.,} ``stick hitting snare"), $\mathcal{A}_{\mathrm{open}}$ is an open anchor vocabulary without predefined categories, and $s_i$ is the confidence score of $a_i$, reflecting how clearly the anchor appears as direct visual evidence of sound generation. Larger $s_i$ in \cref{eq:anchor_def} indicates stronger and more reliable evidence. These anchors from \cref{eq:anchor_voting} serve as fine-grained localization cues that identify specific regions requiring adjustment in Stage 3. \\

\noindent{\textbf{Audio-Visual Consistency.}} This component quantifies the alignment between the predicted localization and the audio-visual evidence to determine refinement necessity. Given the image $I$, audio $A$, initial box $\mathrm{\textit{b}}^{\text{init}}$ from \cref{eq:bbox}, audio label $c_{\mathrm{aud}}$ from \cref{eq:audio_class}, role tags $\mathcal{T}_{\text{role}}$ from \cref{eq:role_tagging}, and anchor evidences $\mathcal{A}_\text{anchor}$ from \cref{eq:anchor_voting}, we define a semantic consistency score:
\begin{equation}
    \mathcal{S}_{\mathrm{av}}
    \;=\;    \mathrm{f}_{\mathrm{con}}(I,\,A,\,\mathrm{\textit{b}}^{\text{init}},\,c_{\mathrm{aud}},\,\mathcal{T}_{\text{role}},\,\mathcal{A}_\text{anchor})
    \in [0,1],
    \label{eq:av_consistency}
\end{equation}
where $\mathcal{S}_{\mathrm{av}}$ is the Audio-Visual Consistency score and $\mathrm{f}_{\mathrm{con}}$ measures how well the predicted box aligns with the semantic evidence inferred from the image and audio, without relying on ground-truth box overlap. Higher scores indicate better alignment between the predicted box and the sound-generating evidence.  \\

\noindent{\textbf{Adaptive Gating.}} This component determines whether refinement is necessary based on multiple quality indicators. We keep the initial box (skip refinement) only when all three conditions are satisfied; otherwise, we perform refinement. The Gating (\textit{G}) decision is defined as:
\begin{equation}
    \mathrm{\textit{G}} = 
    \begin{cases}
    1, & \text{if } (\mathrm{\textit{k}}=1) \;\wedge\; (\mathcal{S}_{\mathrm{av}} \ge \tau_{\mathrm{av}}) \;\wedge\; \
    (s_{\mathrm{aud}} \ge \tau_{\mathrm{aud}})\\
    0, & \text{otherwise}
    \end{cases}
    \label{eq:gating}
\end{equation}
where $\textit{k}$ is a binary keep flag (\textit{k}), with $\mathrm{\textit{k}}=1$ indicating that the initial box is retained and $\mathrm{\textit{k}}=0$ indicating that refinement is required, $\mathcal{S}_{\mathrm{av}}$ is the audio--visual consistency score from \cref{eq:av_consistency} with threshold $\tau_{\mathrm{av}}$, $s_{\mathrm{aud}}$ is the audio confidence score from \cref{eq:audio_class} with threshold $\tau_{\mathrm{aud}}$, and $\wedge$ denotes logical AND. As shown in \cref{eq:gating}, if $\textit{G}=1$, we skip refinement and retain $\mathbf{\textit{b}}^{\text{init}}$; if $\textit{G}=0$, we execute refinement. This adaptive mechanism prevents unnecessary adjustments when the initial prediction is already reliable, improving both efficiency and stability. \\

\noindent{\textbf{Multi-trial Consensus.}}
Since the Analysis stage relies on stochastic decoding, its outputs may vary across runs.
To reduce this variability, we repeat the Analysis stage $n$ times and aggregate the results using the following consensus rules. In our experiments, we set $n{=}5$. ({\textit{i}}) the consistency scores are averaged, ({\textit{ii}}) the top-4 role tags are selected based on their occurrence frequency,
({\textit{iii}}) anchors with identical names are averaged by their confidence scores and only the highest-ranked anchors are retained, and ({\textit{iv}}) the keep flag (\textit{k}) is determined by majority voting. The Audio-Visual Consistency and is computed as:
\begin{equation}
    \bar{\mathcal{S}}_{\mathrm{av}}
    =
    \frac{1}{n}
    \sum_{i=1}^{n}
    \mathcal{S}_{\mathrm{av}}^{(i)},
    \label{eq:consensus_av}
\end{equation}
The final keep decision follows the majority rule defined as:
\begin{equation}
    \textit{k}^{\mathrm{final}}
    =
    \mathbf{1}
    \left(
        \sum_{i=1}^{n} \textit{k}^{(i)} > \frac{n}{2}
    \right).
    \label{eq:consensus_keep}
\end{equation}
\subsection{Stage 3: Refinement}
The Refinement stage aims to correct localization errors identified by the Analysis stage through targeted geometric adjustments. This stage is executed only when Adaptive Gating (\cref{eq:gating}) returns $\mathrm{Gating}\text{ (\textit{G})}=0$. When $\mathrm{Gating}\text{ (\textit{G})}=1$, we skip refinement and retain the initial box:
\begin{equation}
    \textit{G}=1 \;\Rightarrow\; \mathrm{\textit{b}}^{\text{ref}} = \mathrm{\textit{b}}^{\text{init}},
    \label{eq:skip_refinement}
\end{equation}
where $\mathrm{\textit{b}}^{\text{ref}}$ is the final refined bounding box and $\mathrm{\textit{b}}^{\text{init}}$ is the initial box from the Generation stage (\cref{eq:bbox}). Conversely, when Gating $\text{ (\textit{G})}=0$, the stage integrates evidence from the Generation and Analysis stages to produce an improved localization:
\begin{equation}
    \mathrm{\textit{\textit{G}}}=0 \;\Rightarrow\; 
    \mathrm{\textit{b}}^{\text{ref}}=\mathrm{Ref}(I,\; A,\; \mathrm{\textit{b}}^{\text{init}},\; c_{\mathrm{aud}},\; \mathcal{A}_\text{anchor}, \mathcal{T}_{\text{role}}),
    \label{eq:refinement}
\end{equation}
where $\mathrm{Ref}(\cdot)$ in \cref{eq:refinement} is a function that selects and applies geometric operations based on anchor evidences $\mathcal{A}_\text{anchor}$ from \cref{eq:anchor_voting} and role tags $\mathcal{T}_{\text{role}}$ from \cref{eq:role_tagging}. The model adjusts the box through four geometric operations, each designed to address specific types of localization errors.

The model adjusts the box through the operations: \\

\noindent{\textbf{(1) Delta Operation.}}
\begin{equation}
    \mathrm{\textit{b}}^{\text{ref}}
    =
    \begin{bmatrix}
    x_1 + dx + d_\ell & y_1 + dy + d_t\\
    x_2 + dx + d_r & y_2 + dy + d_b
    \end{bmatrix},
    \label{eq:delta_op}
\end{equation}
where \(dx,dy\) shift the whole box toward the confidence-weighted centroid of the anchors from \cref{eq:anchor_voting} that lie outside the current box, and \(d_\ell,d_r,d_t,d_b\) adjust the left/right/top/bottom sides independently. As shown in \cref{eq:delta_op}, this operation is applied when outside anchors indicate a directional bias. \\

\noindent{\textbf{(2) Expand / Shrink Operation.}}
\begin{equation}
    \mathrm{\textit{b}}^{\text{ref}}
    =
    \bigl[
    x_1 - a,\;
    y_1 - a,\;
    x_2 + a,\;
    y_2 + a
    \bigr],
    \label{eq:expand_shrink}
\end{equation}
where \(a>0\) expands and \(a<0\) shrinks the box.
The operation in \cref{eq:expand_shrink} is applied when the center is reasonable but coverage is imbalanced without clear direction,
setting \(a\) based on the outside/total anchor ratio. \\

\begin{table*}[t]
    \renewcommand{\tabcolsep}{2mm}
    \centering
    \caption{Comparison of multi-source sound localization methods on VGGSound-Duet and MUSIC-Duet test sets. We evaluate three types of approaches: (\textit{i}) existing vision-based SSL methods trained with task-specific objectives, (\textit{ii}) off-the-shelf MLLMs baselines (Qwen2.5-Omni, MiniCPM-o, InteractiveOmni) without structured reasoning, and (\textit{iii}) our proposed training-free Generation-Analysis-Refinement framework with $N$ iterations in Stage 2 (Analysis). \best{Bold}/\second{underlined} fonts denote best/second-best performance.}
    \resizebox{0.95\linewidth}{!}
    {
        \begin{tabular}{l  >{\centering\arraybackslash}p{1.8cm} >{\centering\arraybackslash}p{1.8cm} >{\centering\arraybackslash} >{\centering\arraybackslash}p{1.8cm} >{\centering\arraybackslash}p{1.8cm} >{\centering\arraybackslash}p{1.8cm} >{\centering\arraybackslash}p{1.8cm}}
            \Xhline{3\arrayrulewidth}
            
            & \multicolumn{3}{c}{\textbf{VGGSound-Duet} \cite{Vggsound}} & \multicolumn{3}{c}{\textbf{MUSIC-Duet} \cite{The_sound_of_pixels}} \\
            \\[-1em]
            \multicolumn{1}{c}{\multirow{1}{*}[0.7em]{\bf Method}} & \textbf{CAP(\%)} & \textbf{CIoU@0.3(\%)} & \textbf{AUC(\%)} & \textbf{CAP(\%)} & \textbf{CIoU@0.3(\%)} & \textbf{AUC(\%)} \\\hline

            \addlinespace[2pt]
            \multicolumn{7}{c}{\cellcolor{white!20}\textbf{Vision Model}} \\
            \addlinespace[2pt]
            \hline
            \\[-1em]
            Attention\,10k (CVPR'18) \cite{Learning_to_localize_sound_source} & --   & 11.5 & 15.2 & --   & 21.6 & 19.6 \\
            OTS (ECCV'18) \cite{ots_eccv18}            & 10.5 & 12.2 & 15.8 & 11.6 & 13.3 & 18.5 \\
            DMC (CVPR'19) \cite{dmc_cvpr2019}            & --   & 13.8 & 17.1 & --   & 17.5 & 21.1 \\
            CoarseToFIne (ECCV'20) \cite{Multiple_ssl_coarse_fine}            & --   & 14.7 & 18.5 & --   & 17.6 & 20.6 \\
            EZ-VSL (ECCV'22) \cite{localizing_visual_sound_easyway}            & --   & 20.5 & 20.2 & --   & 24.3 & 21.3 \\
            Mix-and-Localize (CVPR'22) \cite{Mix_and_localize}    & 16.3& 21.1 & 20.5 & 47.5 & 26.5 & 21.5 \\
            AVGN (CVPR'23) \cite{Audio_visual_grouping}           & 21.9 & 26.2 & 23.8 & 50.6 & 32.5 & 24.6 \\
            NoPrior (CVPR'24) \cite{NoPrior}        & 32.5 & 46.9 & 29.2 & 52.1 & 38.6 & 30.1 \\
            OA-SSL (CVPR'25) \cite{OA_SSL}         & \second{45.9} & 55.2 & \second{44.8} & \best{61.4} & 45.9 & 36.1 \\

            \addlinespace[4pt]
            \hline
            \addlinespace[4pt]
            
            \multicolumn{7}{c}{\cellcolor{white!10}\textbf{MLLMs}} \\
            \addlinespace[1pt]
            \hline
            \\[-1em]
            Qwen2.5-Omni \cite{qwen25} & 41.0 & 42.6 & 28.3 & 47.2 & 50.6 & 40.8 \\ 
            MiniCPM-o \cite{minicpm} & 36.9 & 38.6 & 26.3 & 29.3 & 27.7 & 23.6 \\ 
            InteractiveOmni \cite{interactiveomni} & 36.0 & 14.6 & 17.9 & 28.8 & 20.0 & 17.0 \\
            \cellcolor{gray!10}\textbf{Ours (N=3)} &
            \cellcolor{gray!10}43.5 &
            \cellcolor{gray!10}\second{59.5} &
            \cellcolor{gray!10}38.2 &
            \cellcolor{gray!10}54.7 &
            \cellcolor{gray!10}\second{80.8} &
            \cellcolor{gray!10}\second{51.4} \\
            \cellcolor{gray!10}\textbf{Ours (N=5)} &
            \cellcolor{gray!10}\best{47.2} &
            \cellcolor{gray!10}\best{77.6} &
            \cellcolor{gray!10}\best{45.8} &
            \cellcolor{gray!10}\second{56.7} &
            \cellcolor{gray!10}\best{82.7} &
            \cellcolor{gray!10}\best{53.2} \\\Xhline{3\arrayrulewidth}
        \end{tabular}
    }
    \vspace{-0.2cm}
    \label{tab:table1}
\end{table*}
\noindent{\textbf{(3) Recenter Operation.}}
\begin{equation}
    \mathrm{\textit{b}}^{\text{ref}}
    =
    \begin{bmatrix}
    c_x^{*} - \tfrac{w}{2} & c_y^{*} - \tfrac{h}{2} \\
    c_x^{*} + \tfrac{w}{2} & c_y^{*} + \tfrac{h}{2}
    \end{bmatrix},
    \label{eq:recenter}
\end{equation}
where \((c_x^{*}, c_y^{*})\) is the target center position (\textit{e.g.,} weighted centroid of outside anchors from \cref{eq:anchor_voting}) and \((w, h)\) are the width and height of the original box. As shown in \cref{eq:recenter}, the refined box maintains the original size \((w, h)\) while shifting the center to \((c_x^{*}, c_y^{*})\). It is applied when the box size is adequate but the center is offset from the sound source.

\section{Experiment}
\label{sec:main_final}
\subsection{Datasets and Evaluation Metrics}
\noindent{\textbf{MUSIC Dataset.}} We evaluate our approach on the MUSIC dataset \cite{The_sound_of_pixels}, which contains 448 real-world YouTube videos featuring musical performances across 11 instrument types in both solo and duet formats. Following the established data splits from prior work \cite{NoPrior, Audio_visual_grouping, T_vsl} to ensure fair comparison, we use the MUSIC-Solo \cite{The_sound_of_pixels} partition (358 training and 90 test) for single-instrument localization and the MUSIC-Duet \cite{The_sound_of_pixels} partition (124 training and 17 test) for multi-instrument scenarios. Our method requires no training data; we simply report results on the designated test sets. \\

\noindent{\textbf{VGG-Sound Dataset.}} The VGG-Sound dataset \cite{Vggsound} encompasses over 200k video clips spanning 221 acoustic categories. For single-source localization, we use the VGG-Sound Source benchmark \cite{Localizing_visual_sounds_hardway} (referred to as VGGSound-Single \cite{Vggsound}). For multi-source evaluation, we follow the protocol from \cite{NoPrior, Audio_visual_grouping, T_vsl}: composite inputs are synthesized by pairing two video frames (448 × 224 resolution) with their synchronized audio signals, and results are reported on the VGGSound-Duet \cite{Vggsound} partition. \\

\noindent{\textbf{Evaluation Metrics.}} Following \cite{Mix_and_localize, NoPrior, Audio_visual_grouping, T_vsl}, we adopt standard evaluation metrics. For single-source localization, we report: Average Precision (AP) measuring the accuracy of the sound source locations, Intersection over Union (IoU) quantifying spatial overlap between predictions and ground-truth, and Area Under the Curve (AUC) evaluating ranking quality across multiple thresholds. For multi-source scenarios, we use Class-aware AP (CAP) and Class-aware IoU (CIoU) to assess per-source localization accuracy and AUC. 
\subsection{Implementation Details}

For each 3-second video, we use the center frame resized to 224×224 and process audio at 16kHz using log-scale mel-spectrograms. Qwen2.5-Omni-7B \cite{qwen25} serves as the backbone MLLM for all stages. The gating mechanism applies fixed thresholds for audio confidence (0.75) and audio–visual consistency (0.5). All experiments are conducted on a single NVIDIA RTX 4090 GPU with consistent settings.

Although our framework requires no training, we report inference cost for completeness. With Qwen2.5-Omni-7B, a single sample requires approximately 4 seconds on average, and the gating mechanism further reduces computation by skipping unnecessary refinement steps.

\begin{table*}[t]
    \renewcommand{\tabcolsep}{2mm}
    \centering
    \caption{Comparison of single-source sound localization methods on VGGSound-Single and MUSIC-Solo test sets. We evaluate three types of approaches: (\textit{i}) existing vision-based SSL methods trained with task-specific objectives, (\textit{ii}) MLLMs baselines (Qwen2.5-Omni, MiniCPM-o, InteractiveOmni) without structured reasoning, and (\textit{iii}) our proposed training-free Generation-Analysis-Refinement framework with $N$ iterations in Stage 2 (Analysis). \best{Bold}/\second{underlined} fonts denote best/second-best performance.}
    \resizebox{0.95\linewidth}{!}
    {%
        \begin{tabular}{l >{\centering\arraybackslash}p{1.8cm} >{\centering\arraybackslash}p{1.8cm} >{\centering\arraybackslash}p{1.8cm} >{\centering\arraybackslash}p{1.8cm} >{\centering\arraybackslash}p{1.8cm} >{\centering\arraybackslash}p{1.8cm}}
            \Xhline{3\arrayrulewidth}
            & \multicolumn{3}{c}{\textbf{VGGSound-Single} \cite{Vggsound}} & \multicolumn{3}{c}{\textbf{MUSIC-Solo} \cite{The_sound_of_pixels}} \\
            
            \multicolumn{1}{c}{\multirow{1}{*}[0.7em]{\bf Method}} & \textbf{AP(\%)} & \textbf{IoU@0.5(\%)} & \textbf{AUC(\%)} & \textbf{AP(\%)} & \textbf{IoU@0.5(\%)} & \textbf{AUC(\%)} \\\hline
            
            \addlinespace[2pt]
            \multicolumn{7}{c}{\cellcolor{white!20}\textbf{Vision Model}} \\
            \addlinespace[2pt]
            \hline
            
            Attention\,10k (CVPR'18) \cite{Learning_to_localize_sound_source} & --   & 19.2 & 30.6   & --   & 37.2 & 38.7 \\
            OTS (ECCV'18) \cite{ots_eccv18}            & 29.8 & 32.8 & 35.7 & 69.3 & 26.1 & 35.8 \\
            DMC (CVPR'19) \cite{dmc_cvpr2019}          & --   & 23.9 & 27.6 & --   & 29.1 & 38.0 \\
            CoarseToFIne (ECCV'20) \cite{Multiple_ssl_coarse_fine} & 28.2 & 29.1 & 34.8 & 70.7 & 33.6 & 39.8 \\
            DSOL (NeurIPS'20) \cite{Discriminative_selfsupervised} & --   & 35.7 & 37.2 & --   & 51.4 & 43.7 \\
            LVS (CVPR'21) \cite{Localizing_visual_sounds_hardway} & 29.6 & 34.4 & 38.2 & 70.6 & 41.9 & 40.3 \\
            EZ-VSL (ECCV'22) \cite{localizing_visual_sound_easyway} & 31.3 & 38.9 & 39.5 & 71.5 & 45.8 & 41.2 \\
            Mix-and-Localize (CVPR'22) \cite{Mix_and_localize} & 32.5 & 36.3 & 38.9 & 68.6 & 30.5 & 40.8 \\
            AVGN (CVPR'23) \cite{Audio_visual_grouping}        & 33.2 & 40.8 & 42.3 & 77.2 & 58.1 & 48.5 \\
            NoPrior (CVPR'24) \cite{NoPrior}           & 46.2 & 41.4 & 41.2 & 77.4 & 62.1 & 59.4 \\
            OA-SSL (CVPR'25) \cite{OA_SSL}             & 51.7 & 47.3 & 44.9 & \second{79.8} & 71.1 & 60.9 \\

            \addlinespace[2pt]
            \hline
            \addlinespace[2pt]
            
            \multicolumn{7}{c}{\cellcolor{white!20}\textbf{MLLM}} \\
            \addlinespace[1.5pt]
            \hline
            
            \\[-1em]
            Qwen2.5-Omni \cite{qwen25}       & 43.6 & 39.4 & 41.8 & 62.7 & 67.8 & 60.7 \\ 
            MiniCPM-o \cite{minicpm}         & 40.9 & 24.9 & 32.1 & 26.2 & 32.8 & 20.1 \\ 
            InteractiveOmni \cite{interactiveomni} & 36.4 & 21.0 & 16.4 & 36.7 & 29.0 & 33.2 \\

            \cellcolor{gray!10}\textbf{Ours (N=3)} &
            \cellcolor{gray!10}\second{60.2} &
            \cellcolor{gray!10}\second{60.1} &
            \cellcolor{gray!10}\second{55.0} &
            \cellcolor{gray!10}78.9 &
            \cellcolor{gray!10}\second{96.2} &
            \cellcolor{gray!10}\second{76.9} \\
            \cellcolor{gray!10}\textbf{Ours (N=5)} &
            \cellcolor{gray!10}\best{60.5} &
            \cellcolor{gray!10}\best{60.2} &
            \cellcolor{gray!10}\best{55.2} &
            \cellcolor{gray!10}\best{80.6} &
            \cellcolor{gray!10}\best{98.5} &
            \cellcolor{gray!10}\best{78.2} \\\Xhline{3\arrayrulewidth}
        \end{tabular}
    }  
    \label{tab:table2}
\end{table*}

\subsection{Comparison to Prior Works}
\noindent{\textbf{Multi-sound Source Localization.}} We compare our method with state-of-the-art methods \cite{Learning_to_localize_sound_source, ots_eccv18, dmc_cvpr2019, Multiple_ssl_coarse_fine, Discriminative_selfsupervised, Localizing_visual_sounds_hardway, localizing_visual_sound_easyway, Mix_and_localize, Audio_visual_grouping, NoPrior, OA_SSL}. As shown in Table \ref{tab:table1}, our method achieves substantial improvements on MUSIC-Duet \cite{The_sound_of_pixels}, outperforming existing methods by 34.9\% in CIoU@0.3 and 15.3\% in AUC. On VGGSound-Duet \cite{Vggsound}, our approach achieves comparable or superior performance, demonstrating enhanced audio-visual scene understanding. \\

\noindent{\textbf{Single-sound Source Localization.}} We conduct comparative experiments on single-source benchmarks against prior methods \cite{Learning_to_localize_sound_source, ots_eccv18, dmc_cvpr2019, Multiple_ssl_coarse_fine, Discriminative_selfsupervised, Localizing_visual_sounds_hardway, localizing_visual_sound_easyway, Mix_and_localize, Audio_visual_grouping, NoPrior, OA_SSL}. Table \ref{tab:table2} reports results on MUSIC-Solo \cite{The_sound_of_pixels} and VGGSound-Single \cite{Vggsound}. On VGGSound-Single \cite{Vggsound}, our method achieves improvements of 8.5\% in AP, 12.8\% in IoU@0.5, and 10.1\% in AUC, with consistent gains on MUSIC-Solo \cite{The_sound_of_pixels}. Overall, our approach matches or surpasses existing methods on both tasks. These results demonstrate that the Generation-Analysis-Refinement (GAR) framework enhances fine-grained audio-visual correspondence through improved scene understanding, enabling more precise sound source localization.

\subsection{Ablation Study}
We quantitatively analyze key design choices in the proposed Generation-Analysis-Refinement pipeline: the number of analysis iterations $N$ in Analysis stage, the contribution of each stage, and comparison with existing methods. \\

\noindent{\textbf{Effect of Analysis Iterations.}} We evaluate the impact of analysis iterations $N\in\{1,3,5\}$ on single- and multi-source benchmarks. As shown in Table \ref{tab:table3_n_iteration_solo} and Table \ref{tab:table4_n_iteration_duet}, increasing $N$ consistently improves performance across all metrics, with $N{=}5$ achieving the best results. This demonstrates that iterative refinement effectively corrects localization errors in single-source scenarios and enhances source discrimination in complex multi-source scenarios. \\

\begin{table}[t]
  \centering
  \vspace{-0.3cm}
  \caption{Effect of the number of analysis iterations ($N$) in Stage 2. The Stage 2 is repeated $N$ times per sample, with multi-trial outputs aggregated through statistical consensus to enhance stability. Results on VGGSound-Single and MUSIC-Solo for $N\in\{1, 3, 5\}$.}
  \label{tab:table3_n_iteration_solo}
  \setlength{\tabcolsep}{2pt}
  \renewcommand{\tabcolsep}{2.5mm}

  \begin{adjustbox}{max width=\columnwidth}
  \begin{tabular}{ccccccc}
    \Xhline{3\arrayrulewidth}
    \multicolumn{1}{c}{\multirow{2}{*}[-0.1em]{\textbf{N}}} &
    \multicolumn{3}{c}{\textbf{VGGSound-Single}} &
    \multicolumn{3}{c}{\textbf{MUSIC-Solo}} \\
    \cmidrule(lr){2-4}\cmidrule(l){5-7}
     & \textbf{AP} & \textbf{IoU@0.5} & \textbf{AUC} &
       \textbf{AP} & \textbf{IoU@0.5} & \textbf{AUC} \\
    \midrule
    1 & 60.1 & 60.0 & 55.1 & 78.8 & 96.3 & 76.8 \\
    3 & 60.2 & 60.1 & 55.0 & 78.9 & 96.2 & 76.9 \\
    \cellcolor{gray!10}5 & 
    \cellcolor{gray!10}\textbf{60.5} & 
    \cellcolor{gray!10}\textbf{60.2} & 
    \cellcolor{gray!10}\textbf{55.2} &
    \cellcolor{gray!10}\textbf{80.6} & 
    \cellcolor{gray!10}\textbf{98.5} & 
    \cellcolor{gray!10}\textbf{78.2} \\
    \Xhline{3\arrayrulewidth}
  \end{tabular}
  \end{adjustbox}
\end{table}
\begin{table}[t]
  \centering
  \vspace{-0.15cm}
  \caption{Effect of the number of analysis iterations ($N$) in Stage 2. The Stage 2 is repeated $N$ times per sample, with multi-trial outputs aggregated through statistical consensus to enhance stability. Results on VGGSound-Duet and MUSIC-Duet for $N\in\{1, 3, 5\}$.}
  \label{tab:table4_n_iteration_duet}
  \setlength{\tabcolsep}{1pt}
  \renewcommand{\tabcolsep}{2.0mm}
  \begin{adjustbox}{max width=\columnwidth}
  \begin{tabular}{ccccccc}
    \Xhline{3\arrayrulewidth}
    \multicolumn{1}{c}{\multirow{2}{*}[-0.1em]{\textbf{N}}} &
    \multicolumn{3}{c}{\textbf{VGGSound-Duet}} &
    \multicolumn{3}{c}{\textbf{MUSIC-Duet}} \\
    \cmidrule(lr){2-4}\cmidrule(l){5-7}
     & \textbf{CAP} & \textbf{CIoU@0.3} & \textbf{AUC} &
       \textbf{CAP} & \textbf{CIoU@0.3} & \textbf{AUC} \\
    \midrule
    1 & 43.4 & 58.9 & 38.1 & 54.5 & 80.7 & 51.5 \\
    3 & 43.5 & 59.5 & 38.2 & 54.7 & 80.8 & 51.4 \\
    \cellcolor{gray!10}5 & 
    \cellcolor{gray!10}\textbf{47.2} & 
    \cellcolor{gray!10}\textbf{77.6} & 
    \cellcolor{gray!10}\textbf{45.8} &
    \cellcolor{gray!10}\textbf{56.7} & 
    \cellcolor{gray!10}\textbf{82.7} & 
    \cellcolor{gray!10}\textbf{53.2} \\
    \Xhline{3\arrayrulewidth}
  \end{tabular}
  \end{adjustbox}
\end{table}
\begin{figure*}
    \centering
    \includegraphics[width=0.9\linewidth]{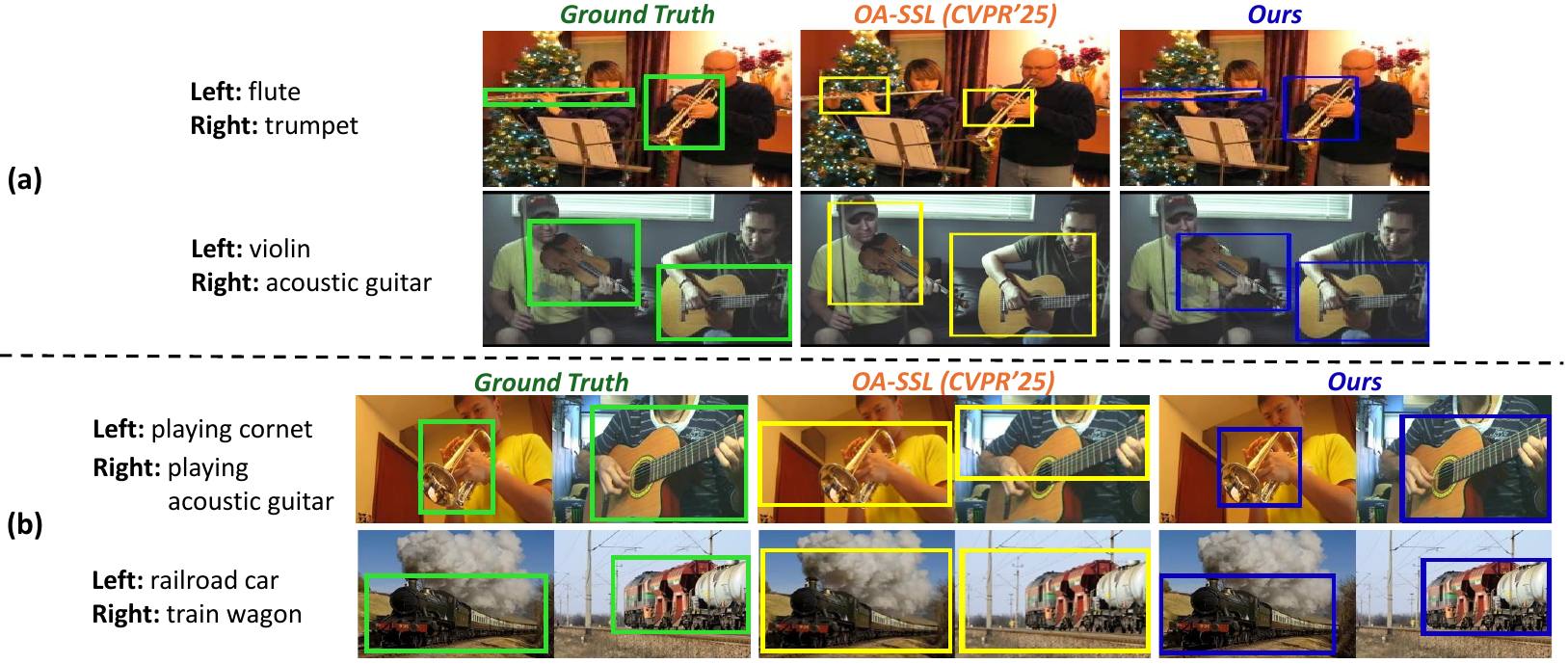}
    \vspace{-0.2cm}
    \caption{Visualization results for (a) MUSIC-Duet and (b) VGGSound-Duet test set. We compare our method with OA-SSL\cite{OA_SSL}. More comparisons are in the supplementary document.}
    \vspace{-0.5cm}
    \label{fig:figure3}
\end{figure*}
\begin{figure}
    \centering
    \includegraphics[width=\linewidth]{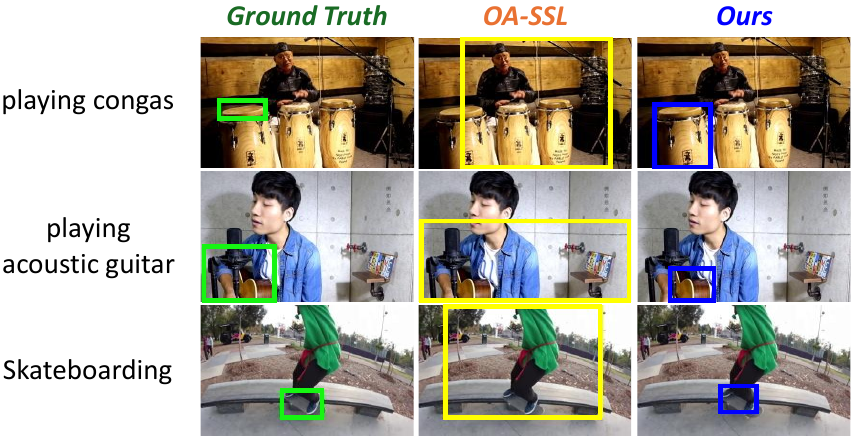}
    \vspace{-0.2cm}
    \caption{Visualization results for VGGSound-Single test set. We compare our method with OA-SSL \cite{OA_SSL}. More comparisons are in the supplementary document.}
    \vspace{-0.2cm}
    \label{fig:figure4}
\end{figure}
\noindent{\textbf{Effect of Each Stage.}} Table \ref{tab:table5_ablation} shows the effect of stage combinations on VGGSound-Duet. Using only Stage 1 (Generation) provides baseline performance, while activating all stages leads to substantial improvements across all metrics. Notably, CIoU@0.3 increases by 16.9 percentage points, attributed to Stage 2 (Analysis) iterative analysis enhancing candidate box consistency and Stage 3 (Refinement) making fine-grained adjustments. Stage 2 (Analysis) drives a gating mechanism that determines whether Stage 3 is executed. It enables the framework to skip unnecessary refinement and perform fine-grained adjustments only when needed. \\

\noindent{\textbf{Evaluation with Different MLLMs.}} Table \ref{tab:table6_mllm} summarizes the performance of the Generation-Analysis-Refinement (GAR) framework on VGGSound-Duet~\cite{Vggsound} with different MLLMs (Qwen2.5-Omni-3B/7B~\cite{qwen25}). The 7B model achieves stronger performance across all metrics, demonstrating that a more capable MLLM improves localization accuracy in multi-source scenarios.

\begin{table}[t!]
  \centering
  \vspace{-0.4cm}
  \caption{Effect of the proposed method on VGGSound-Duet. Stage 2 (Analysis) and Stage 3 (Refinement) are evaluated together because the gating mechanism in Stage 2 determines whether Stage 3 should be executed, making them functionally interdependent.}
  \label{tab:table5_ablation}

  \setlength{\tabcolsep}{2pt}
  \renewcommand{\arraystretch}{0.95}
  \renewcommand{\tabcolsep}{1mm}
  
  \begin{adjustbox}{max width=\columnwidth}
  \begin{tabular}{cccccc}
    \Xhline{3\arrayrulewidth}
    \\[-0.8em]
    \textbf{Stage 1} & \textbf{Stage 2} & \textbf{Stage 3} &
    \textbf{CAP (\%)} & \textbf{CIoU@0.3 (\%)} & \textbf{AUC (\%)} \\
    \midrule
    \\[-0.8em]
    \cmark & --         & --         & 41.0 & 42.6 & 28.3 \\
    \\[-0.8em]
    \cellcolor{gray!10}\cmark & \cellcolor{gray!10}\cmark & \cellcolor{gray!10}\cmark & \cellcolor{gray!10}\textbf{43.5} & \cellcolor{gray!10}\textbf{59.5} & \cellcolor{gray!10}\textbf{38.2} \\
    \\[-0.8em]
    \Xhline{3\arrayrulewidth}
  \end{tabular}
  \end{adjustbox}
    \vspace{-0.3cm}
\end{table}
\subsection{Visualization Results}
Figures \ref{fig:figure3} and \ref{fig:figure4} show qualitative comparisons in single-source and multi-source settings. Our method more accurately isolates true sound-emitting objects compared to OA-SSL \cite{OA_SSL} and avoids incorrect regions, demonstrating improved spatial precision. These results visually confirm the effectiveness of the proposed three-stage framework.

\begin{table}[t]
  \centering
  \vspace{-0.1cm}
  \caption{Comparison of different MLLM backbones (Qwen2.5-Omni-3B \textit{vs.} 7B) in the proposed framework on VGGSound-Duet. Both models serve as the foundation for all three stages (Generation, Analysis, Refinement) with analysis iterations fixed at $N=3$.}
  \label{tab:table6_mllm}

  \setlength{\tabcolsep}{2pt}
  \renewcommand{\arraystretch}{0.95}
  \renewcommand{\tabcolsep}{1mm}

  \begin{adjustbox}{max width=\columnwidth}
  \begin{tabular}{lccc}
    \Xhline{3\arrayrulewidth}
    \\[-0.8em]
    \textbf{Model} & \textbf{CAP(\%)} & \textbf{CIoU@0.3(\%)} & \textbf{AUC(\%)} \\
    \midrule
    Qwen2.5-Omni-3B \cite{qwen25} & 39.9 & 49.8 & 33.0 \\ 
    \\[-0.8em]
    \cellcolor{gray!10}Qwen2.5-Omni-7B \cite{qwen25}& \cellcolor{gray!10}\best{43.5} & \cellcolor{gray!10}\best{59.5} & \cellcolor{gray!10}\best{38.2} \\
    \Xhline{3\arrayrulewidth}
  \end{tabular}
  \end{adjustbox}
\end{table}

\subsection{Discussion}
Our study demonstrates that strong SSL performance can be achieved without task-specific training by leveraging the inherent reasoning capabilities of MLLMs. The proposed role tagging, anchor voting, and adaptive gating contribute to both interpretability and efficiency. However, iterative analysis increases inference time, and performance depends on the underlying MLLMs. Future work will focus on reducing computational cost, incorporating temporal reasoning, and validating generalization to broader real-world scenarios.

\section{Conclusion}
We presented a training-free audio-visual sound source localization framework based on a Generate-Analyze-Refine pipeline with MLLMs. By reformulating SSL as a cognitive reasoning process, the method achieved competitive performance on both single-source and multi-source benchmarks. Open-set role tagging and anchor voting provided interpretable spatial confidence, while adaptive gating enabled efficient refinement. These results highlight the potential of pre-trained MLLMs for fine-grained audio-visual correspondence and complex multimodal perception tasks.

\section*{Acknowledgements}
This work was partly supported by IITP-ITRC grant funded by the Korea government (MSIT)(IITP-2026-RS-2023-00258649, 40\%) and partly supported by IITP grant funded by the Korea government (MSIT)(No. RS-2022-II220124, Development of Artificial Intelligence Technology for Self-Improving Competency-Aware Learning Capabilities (30\%), No. RS-2024-00509257: Global AI Frontier Lab (30\%)).

{
\small
\bibliographystyle{ieeenat_fullname}
\bibliography{main}
}


\definecolor{cvprblue}{rgb}{0.21,0.49,0.74}
\hypersetup{
    pagebackref,
    breaklinks,
    colorlinks,
    citecolor=cvprblue
}

\def\maketitlesupplementary
   {
   \newpage
       \twocolumn[
        \centering
        \Large
        \textbf{\thetitle\\--\textit{Supplementary Material}--}\\
        \vspace{1.5em}
       ] 
   }

\newtcolorbox{mybox}[1][]{%
    enhanced,
    top=1pt,
    bottom=1pt,
    boxrule=0pt,
    arc=0pt,
    colframe=white,
    colback=white,
    borderline north={0.5mm}{0mm}{black},
    borderline south={0.5mm}{0mm}{black},
    #1
}

\newtcolorbox{mybox2}[1][]{%
    enhanced,
    top=1pt,
    bottom=1pt,
    boxrule=0pt,
    arc=0pt,
    colframe=white,
    colback=white,
    borderline north={0.5mm}{0mm}{black},
    borderline south={0.5mm}{0mm}{black},
    width=\textwidth, 
    left=0pt,
    right=0pt,
    #1
}

\maketitlesupplementary

\renewcommand{\thefigure}{S.\arabic{figure}}
\renewcommand{\thetable}{S.\arabic{table}}
\renewcommand{\theequation}{S.\arabic{equation}}

\makeatletter
\renewcommand{\@biblabel}[1]{[S#1]}
\makeatother 

\noindent This supplementary material provides additional implementation details and extended experimental results for the proposed method. First, it presents experimental results analyzing the impact of threshold adjustments at each stage, with the number of iterations in the Analysis stage fixed at 5, based on the VGGSound dataset. Next, it shows comparative results between various prompt variations and the proposed method, and further explains the effectiveness of the approach through additional visualization materials. Finally, it discloses the detailed prompts used throughout the entire Generate-Analysis-Refinement process.
\vspace{-0.1cm}

\begin{tcolorbox}[
    colback=gray!10,
    colframe=gray!60,
    boxrule=0.8pt,
    arc=2mm,
    left=4mm, right=4mm, top=2mm, bottom=2mm,
    title=Table of Contents
]
\begin{enumerate}[nosep, leftmargin=*]
    \item Additional Experimental Results
    \item Prompt Variation Comparison
    \item Additional Visualization Results 
    \item Prompts for Proposed Method
\end{enumerate}
\end{tcolorbox}
\vspace{-0.2cm}

\section{Additional Experimental Results}
We analyzed the impact of the Audio Confidence threshold ($A_C$) in Stage 1 (Generation) with the number of iterations fixed at $N=5$.
Table \ref{tab:table1_supple} presents results on the VGGSound-Single [\textcolor{cyan}{P16}] dataset. Performance remains stable despite variations in the audio confidence ($A_C$) and Audio-Visual Consistency ($AV_C$) thresholds, with the best performance achieved at $A_C{=}0.75$ and $AV_C{=}0.5$. For single-sound source datasets VGGSound-Single, while the SOTA performance is AP 51.7, IoU@0.5 47.3, and AUC 44.9, our proposed method significantly surpasses these with AP 60.5, IoU@0.5 60.2, and AUC 55.2. These results demonstrate that the proposed method is robust to threshold variations while consistently maintaining strong performance on the VGGSound-Single dataset.

\begin{table}[t]
  \centering
  \caption{Analysis of the impact of Audio Confidence ($A_C$) and Audio-Visual Consistency ($AV_C$) thresholds on performance. The number of iterations in the Stage 2 (Analysis) is fixed at $N=5$. Evaluated on the VGGSound-Single dataset.}
  \label{tab:table1_supple}
  \setlength{\tabcolsep}{1pt}
  \renewcommand{\tabcolsep}{1.5mm}
  \begin{adjustbox}{max width=\columnwidth}
  \begin{tabular}{cc ccc}
    \Xhline{3\arrayrulewidth}
    \multicolumn{2}{c}{} &
    \multicolumn{3}{c}{\textbf{VGGSound-Single} [\textcolor{cyan}{P16}]} \\
    \cmidrule(lr){3-5}
    \textbf{$A_C$} & \textbf{$AV_C$} &
    \textbf{AP} & \textbf{IoU@0.5} & \textbf{AUC} \\
    \midrule
    0.5  & 0.5  & 60.1 & 60.0 & 55.0 \\
    0.75 & 0.5  & \textbf{60.5} & \textbf{60.2} & \textbf{55.2} \\
    0.5  & 0.75 & 60.1 & 60.0 & 55.0 \\
    0.75 & 0.75 & 60.2 & 60.1 & 55.1 \\
    \Xhline{3\arrayrulewidth}
  \end{tabular}
  \end{adjustbox}
\end{table}

\section{Prompt Variation Comparison}
In this experiment, we design four different prompt-based methods that apply varying conditions and constraints to perform Sound Source Localization in a more fine-grained. \\

\noindent\textbf{Method 1 (Direct Estimation):} This method represents the simplest approach, directly generating multiple candidate bounding boxes from the image and audio. The generated candidates are self-examined to assess the appropriateness of bounding box sizes and identify positional errors, with suggestions for improvements. Finally, based on the inspection results, the bounding boxes are refined to select the optimal candidate. When refinement is needed, a conservative rule of adjusting by at least 1 pixel is applied, serving as a basic calibration that quickly validates the initial box. \\

\noindent\textbf{Method 2 (Class-Conditional Refinement):} This method applies stronger structural constraints than the Method 1. First, an initial bounding box is estimated from the image and audio, and separately, the audio source class (\textit{e.g.,} ``violin", ``dog barking") is extracted using only the audio. Subsequently, refinement is performed by considering both the initial bounding box and the extracted audio class together, ensuring that the bounding box logically aligns with the audio source class. \\

\noindent\textbf{Method 3 (Anchor-Guided Refinement):} This method extends the Method 2 by providing more detailed analysis information. Beyond the audio class and initial bounding box, it explicitly identifies visual sub-parts (anchors) that generate the sound source. For example, in the case of a violin, anchors such as ``bow-string contact point" and ``violin body" are identified. The model analytically interprets the relationships among the audio class, initial bounding box, and visible anchors to perform refinement. This method focuses on identifying and utilizing fine-grained parts of the sound source. \\

\noindent\textbf{Our Method (Generation-Analysis-Refinement):} The final our method extends the Method 3 and represents the final approach proposed in this paper. In this method, all meta-analysis information including Audio-Visual Consistency ($AV_C$), role tags, and anchor votes is provided as input, designed to enable the model to comprehensively verify judgments from previous stages. Additionally, we analyze the progressive improvement effect by adjusting the number of iterations N in the refinement stage ($N{=}1,3,5$). Through this, we systematically compare the performance of each method and demonstrate the superiority of the proposed approach. \\

Using the above-mentioned four methods described above, we compare performance across single-sound and multi-sound source settings. Table \ref{tab:table3_supple} and Table \ref{tab:table4_supple} summarize the prompt variation experiment results for single-sound source and multi-sound source datasets. Ours ($N=5$) demonstrates the best overall performance, achieving 60.5\% AP on VGGSound-Single [\textcolor{cyan}{P16}], 80.6\% AP on MUSIC-Solo [\textcolor{cyan}{P13}], 47.2\% CAP on VGGSound-Duet [\textcolor{cyan}{P16}], and 56.7\% CAP on MUSIC-Duet. Performance progressively improves from Method 1 to the proposed method, and also consistently enhances as the number of iterations \textit{N} increases. This clearly confirms the effectiveness of integrating meta-analysis information and iterative refinement.

\begin{table}[t]
  \centering
  \caption{Performance comparison of various prompt variation methods on single-sound source datasets. Method 1 performs basic refinement with minimal adjustments. Method 2 incorporates audio class information for refinement. Method 3 leverages detailed analysis information including visual anchors. Ours represents the proposed meta-analysis-based method with varying iteration counts ($N{=}1, 3, 5$). Evaluated on VGGSound-Single and MUSIC-Solo datasets using CAP, CIoU@0.3, and AUC metrics.}
  \label{tab:table3_supple}
  \setlength{\tabcolsep}{1pt}
  \renewcommand{\tabcolsep}{1.5mm}
  \begin{adjustbox}{max width=\columnwidth}
  \begin{tabular}{ccccccc}
    \Xhline{3\arrayrulewidth}
    \multicolumn{1}{c}{\multirow{2}{*}[-0.1em]{\textbf{Method}}} &
    \multicolumn{3}{c}{\textbf{VGGSound-Single} [\textcolor{cyan}{P16}]} &
    \multicolumn{3}{c}{\textbf{MUSIC-Solo} [\textcolor{cyan}{P13}]} \\
    \cmidrule(lr){2-4}\cmidrule(l){5-7}
     & \textbf{AP} & \textbf{IoU@0.5} & \textbf{AUC} &
       \textbf{AP} & \textbf{IoU@0.5} & \textbf{AUC} \\
    \midrule
    Method 1 & 52.0 & 46.5 & 44.5 & 81.4 & 96.5 & 78.8 \\
    Method 2 & 59.5 & 59.0 & 54.2 & 82.7 & 98.9 & 80.2 \\
    Method 3 & 60.0 & 59.7 & 54.9 & 81.6 & 97.6 & 79.1 \\
    Ours ($N=1$) & 60.1 & 60.0 & 55.0 & 78.8 & 96.3 & 76.8 \\
    Ours ($N=3$) & 60.2 & 60.1 & 55.0 & 78.9 & 96.2 & 76.9 \\
    \textbf{Ours ($N=5$)} & \textbf{60.5} & \textbf{60.2} & \textbf{55.2} & \textbf{80.6} & \textbf{98.5} & \textbf{78.2} \\
    \Xhline{3\arrayrulewidth}
  \end{tabular}
  \end{adjustbox}
\end{table}

\begin{table}[t]
  \centering
  \caption{Performance comparison of various prompt variation methods on multi-sound source datasets. Method 1 performs basic refinement with minimal adjustments. Method 2 incorporates audio class information for refinement. Method 3 leverages detailed analysis information including visual anchors. Ours represents the proposed meta-analysis-based method with varying iteration counts ($N{=}1, 3, 5$). Evaluated on VGGSound-Duet and MUSIC-Duet datasets using CAP, CIoU@0.3, and AUC metrics.}
  \label{tab:table4_supple}
  \setlength{\tabcolsep}{1pt}
  \renewcommand{\tabcolsep}{1.5mm}
  \begin{adjustbox}{max width=\columnwidth}
  \begin{tabular}{ccccccc}
    \Xhline{3\arrayrulewidth}
    \multicolumn{1}{c}{\multirow{2}{*}[-0.1em]{\textbf{Method}}} &
    \multicolumn{3}{c}{\textbf{VGGSound-Duet} [\textcolor{cyan}{P16}]} &
    \multicolumn{3}{c}{\textbf{MUSIC-Duet} [\textcolor{cyan}{P13}]} \\
    \cmidrule(lr){2-4}\cmidrule(l){5-7}
     & \textbf{CAP} & \textbf{CIoU@0.3} & \textbf{AUC} &
       \textbf{CAP} & \textbf{CIoU@0.3} & \textbf{AUC} \\
    \midrule
    Method 1 & 44.7 & 57.0 & 37.7 & 46.5 & 77.9 & 45.1 \\
    Method 2 & 32.9 & 23.0 & 26.5 & 44.7 & 36.1 & 44.6 \\
    Method 3 & 45.5 & 60.4 & 39.5 & 53.6 & 76.9 & 49.4 \\
    Ours ($N=1$) & 43.4 & 58.9 & 38.1 & 54.7 & 80.8 & 51.4 \\
    Ours ($N=3$) & 43.5 & 59.5 & 38.2 & 54.7 & 80.8 & 51.4 \\
    \textbf{Ours ($N=5$)} & \textbf{47.2} & \textbf{77.6} & \textbf{45.8} & \textbf{56.7} & \textbf{82.7} & \textbf{53.2} \\
    \Xhline{3\arrayrulewidth}
  \end{tabular}
  \end{adjustbox}
\end{table}

\section{Additional Visualization Results}
\begin{figure}
    \centering
    \includegraphics[width=1.0\linewidth]{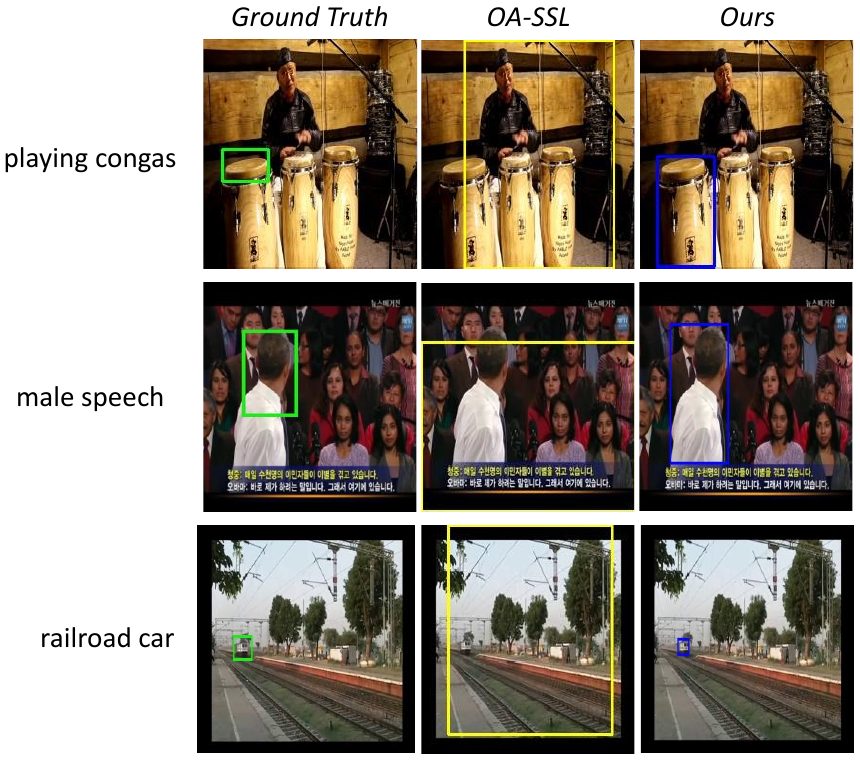}
    \caption{Visualization of sound source localization results in VGGSound-Single [\textcolor{cyan}{P16}] dataset. Each row represents a different example, and each column shows the original image, Ground Truth (actual sound source location), prediction results of the OA-SSL [\textcolor{cyan}{P38}] method, and prediction results of the proposed method (Ours).}
    \label{fig:figure1_supple}
\end{figure}

\begin{figure}
    \centering
    \includegraphics[width=1.0\linewidth]{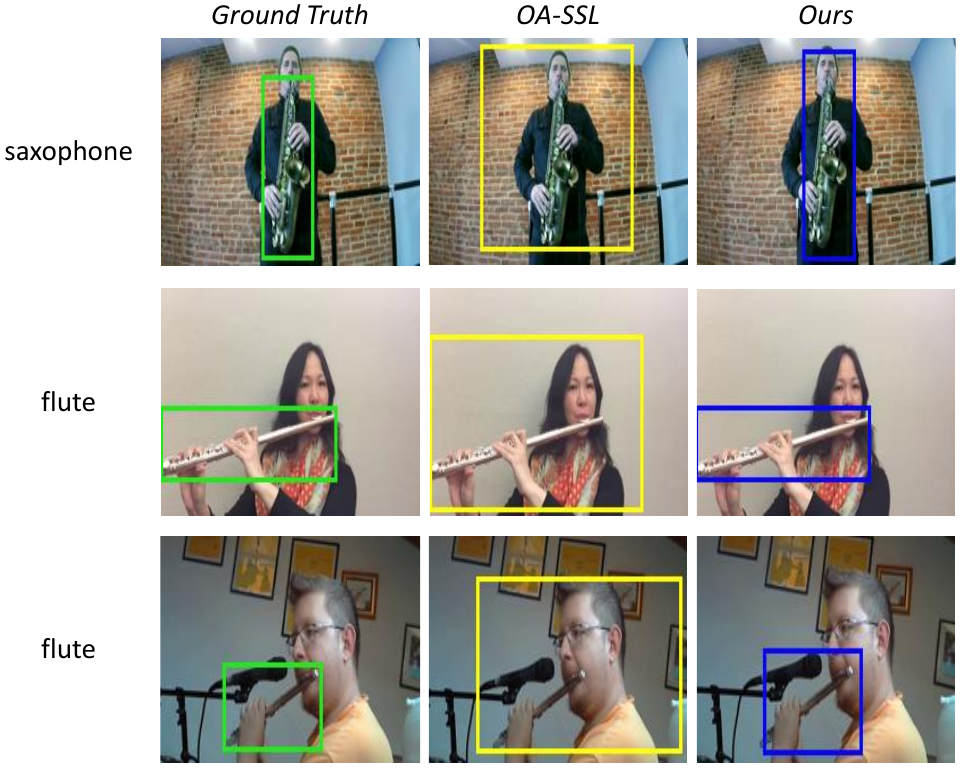}
    \caption{Visualization of sound source localization results on the MUSIC-Solo [\textcolor{cyan}{P13}] dataset. Each row shows a different instrument example. From left to right, we present the Ground Truth bounding box, the prediction produced by OA-SSL [\textcolor{cyan}{P38}], and the prediction generated by our proposed method (Ours).}
    \label{fig:figure2_supple}
\end{figure}

\begin{figure*}[t]
    \centering
    \includegraphics[width=0.95\linewidth]{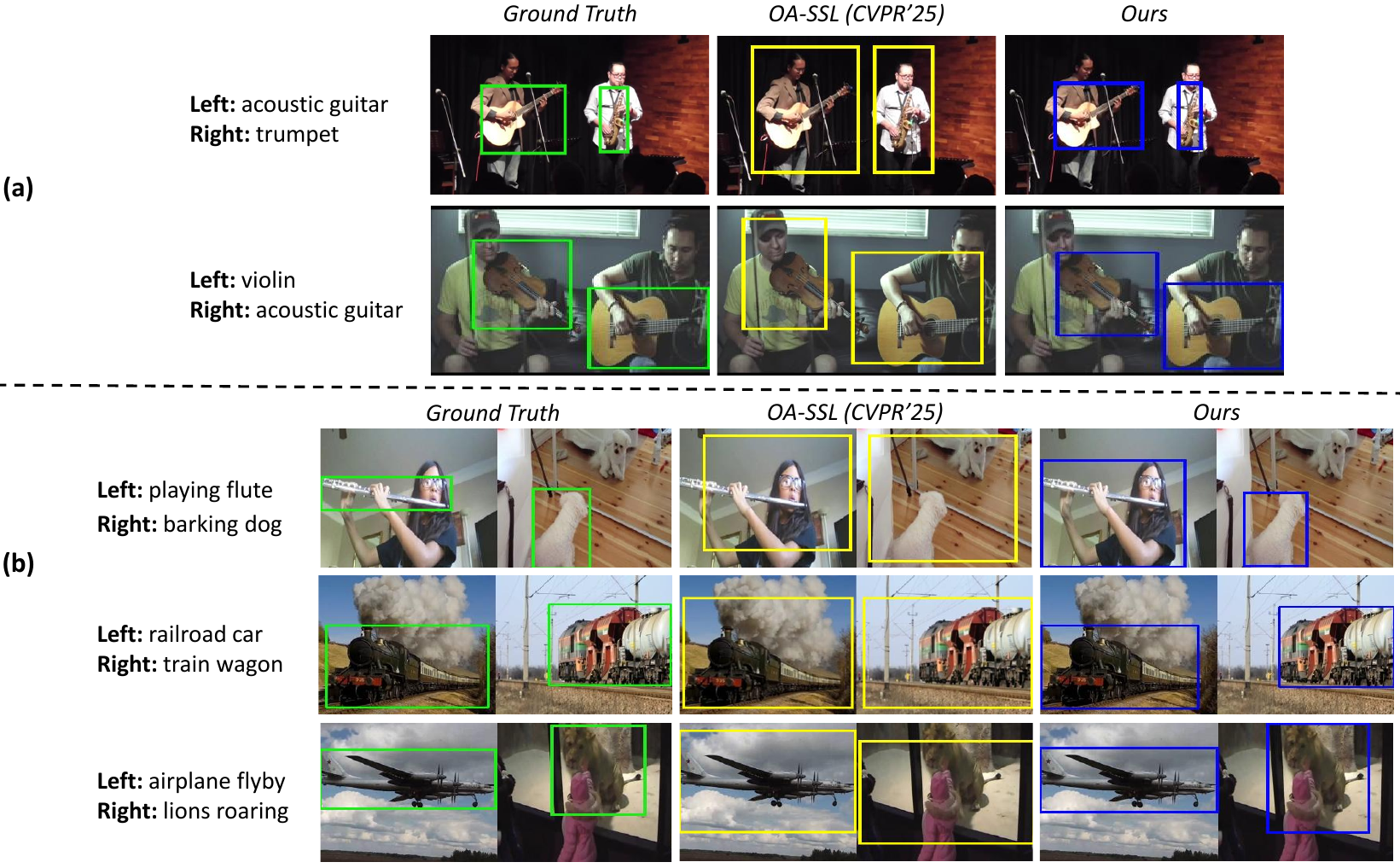}
    \caption{Visualization of sound source localization results in VGGSound-Duet [\textcolor{cyan}{P16}] and MUSIC-Duet [\textcolor{cyan}{P13}] datasets. \textbf{(a)} Results of simultaneously localizing two instrumental sound sources in VGGSound-Duet [\textcolor{cyan}{P16}]. \textbf{(b)} Results in more complex multi-sound source environments in MUSIC-Duet [\textcolor{cyan}{P13}]. Each example includes the original image, Ground Truth, OA-SSL [\textcolor{cyan}{P38}] prediction, and the proposed method prediction (Ours).}
    \label{fig:figure3_supple}
\end{figure*}

Figures \ref{fig:figure1_supple}, \ref{fig:figure2_supple}, \ref{fig:figure3_supple} visually compare the sound source localization results of the proposed method and the existing method (OA-SSL [\textcolor{cyan}{P38}]). In each example, Ground Truth represents the actual location of the sound source, OA-SSL shows the prediction results of the existing method, and Ours indicates the results of the proposed method.

Figure \ref{fig:figure1_supple} shows single-source results, where our method consistently produces tighter and more correctly positioned bounding boxes than OA-SSL [\textcolor{cyan}{P38}] across all examples. Figure \ref{fig:figure2_supple} further demonstrates improved precision on MUSIC-Solo [\textcolor{cyan}{P13}] with saxophone and flute cases, where our method more accurately aligns with the actual sound-producing regions. Figure \ref{fig:figure3_supple}(a) illustrates multi-source scenarios involving two instruments. While OA-SSL struggles with scale and placement, our approach more clearly separates and localizes each source. Figure \ref{fig:figure3_supple}(b) presents more challenging multi-source scenes with visually separated sources. Our method maintains accurate and compact localization, whereas OA-SSL often generates overly large regions. Overall, our approach yields consistently tighter and more reliable localization than OA-SSL across both single-source and multi-source settings.

\section{Prompts for Proposed Method}
In this study, we design a structured prompt framework to process visual and audio information in a step-by-step manner. Stage 1 (Generation) consists of two sub-stages: Stage 1 (Generation): Audio-Visual Localization Table \ref{tab:prompt_stage1-1} estimates the location of the primary sound source object by utilizing both image and audio, while Stage 1 (Generation): Audio Classification Table \ref{tab:prompt_stage1-2} classifies sound events based solely on audio. Stage 2 (Analysis) Table \ref{tab:prompt_stage2} verifies whether the predicted sound source actually matches visually based on the information generated in Stage 1 (Generation), and quantifies this as Audio-Visual Consistency ($AV_C$). Finally, Stage 3 (Refinement) Table \ref{tab:prompt_stage3} refines the bounding box based on these analysis results, achieving meaningful improvements with minimal changes. These three stages of the prompt framework are combined to enable step-by-step and interpretable single-sound source and multi-sound source localization without training.

\section*{References}

\small [P13] Zhao Hang, Gan Chuang, Rouditchenko Andrew, Vondrick Carl, McDermott Josh, and Torralba Antonio. The sound of pixels. In \textit{ECCV}, 2018. \\
\small [P16] Chen Honglie, Xie Weidi, Vedaldi Andrea, and Zisserman Andrew. Vggsound: A large-scale audio-visual dataset. In \textit{ICASSP}, 2020. \\
\small [P38] Sung Jin Um, Dongjin Kim, Sangmin Lee, and Jung Uk Kim. In \textit{CVPR}, 2025. \\

\clearpage
\onecolumn

\begin{table}[h]
    \centering
    \caption{Stage 1 (Generation) Prompt: Audio-Visual Localization}
    \label{tab:prompt_stage1-1}
    \begin{tabular}{p{0.95\textwidth}}
    \toprule
    \textbf{Prompt:} \\
    \midrule
    You are an assistant for audio-visual sound source localization (SSL). \\
    \\
    \textbf{TASK (Stage A):} \\
    Given an IMAGE and an AUDIO clip from the same scene: \\
    1) Locate exactly one \emph{main} sound-emitting object in the image and output its bounding box as \([x1, y1, x2, y2]\). \\
    2) Provide a concise visual description of the sound-emitting object. \\
    \\
    \textbf{STRICT OUTPUT:} \\
    \begin{minipage}{\linewidth}
    \begin{verbatim}
    {
      "bbox": [x1, y1, x2, y2],
      "description": "visual description of the 
                      sound-emitting object"
    }
    \end{verbatim}
    \end{minipage} \\
    \\
    - The bbox must be four integers in the original image coordinates (x1<x2, y1<y2). \\
    - Do not output any text or fields outside the JSON object. \\
    \bottomrule
    \end{tabular}
\end{table}

\vspace{2em}

\begin{table}[h]
    \centering
    \caption{Stage 1 (Generation) Prompt: Audio Classification}
    \label{tab:prompt_stage1-2}
    \begin{tabular}{p{0.95\textwidth}}
    \toprule
    \textbf{Prompt:} \\
    \midrule
    You are an audio classification expert. \\
    \\
    \textbf{TASK (Stage B):} \\
    Listen to the AUDIO and classify the dominant audio event using a short, lowercase class name \\
    (e.g., ``violin", ``piano", ``dog barking", ``engine", ``drum set"). \\
    You must also provide a confidence score in the range \([0.0, 1.0]\). \\
    \\
    \textbf{STRICT OUTPUT:} \\
    \begin{minipage}{\linewidth}
    \begin{verbatim}
    {
      "audio_class": "<concise class name>",
      "audio_confidence_score": <float>
    }
    \end{verbatim}
    \end{minipage} \\
    \\
    - The class name must be lowercase and concise. \\
    - The confidence must be a float between 0.0 and 1.0. \\
    - Do not include any text outside the JSON. \\
    \bottomrule
    \end{tabular}
\end{table}

\clearpage
\twocolumn

\clearpage
\onecolumn

\begin{table}[h]
    \centering
    \caption{Stage 2 (Analysis) Prompt}
    \label{tab:prompt_stage2}
    \begin{tabular}{p{0.95\textwidth}}
    \toprule
    \textbf{Prompt:} \\
    \midrule
    You must verify whether the sound suggested by the AUDIO is actually \emph{visibly supported} within the IMAGE. \\
    You must rely only on the given image–audio pair and must not hallucinate unseen content. \\
    \\
    \textbf{Context:} \\
    - previous\_bbox \\
    - audio\_class \\
    - audio\_confidence\_score \\
    - image size \([W \times H]\) \\
    \\
    \textbf{Definitions:} \\
    \\
    - \textbf{anchor\_votes}: propose 0–5 concise, lowercase visual anchors that represent visible causes of the sound indicated by the audio class. \\
    \quad Examples: \\
    \quad - applause → ``hands\_clapping" \\
    \quad - violin → ``bow\_on\_strings", ``violin\_body" \\
    \quad - dog barking → ``dog\_mouth\_open" \\
    \\
    \quad Format: \\
    \begin{minipage}{\linewidth}
    \begin{verbatim}
    {"anchor":"<token_with_underscores>", "score": s}
    \end{verbatim}
    \end{minipage}
    \quad where \(s \in [0,1]\). \\
    \\
    - \textbf{role\_tags}: up to four short tokens summarizing the visual roles or cues relied upon. \\
    - \textbf{av\_consistency}: audio–visual consistency score \([0,1]\), based on \\
    \quad (i) alignment between audio class and visible evidence, \\
    \quad (ii) spatial proximity to previous bbox, \\
    \quad (iii) clarity of the visible cues. \\
    - \textbf{keep}: true only when refinement can be safely skipped. \\
    \\
    \textbf{STRICT OUTPUT:} \\
    \begin{minipage}{\linewidth}
    \begin{verbatim}
    {
      "av_consistency": <float>,
      "role_tags": [...],
      "anchor_votes": [...],
      "keep": <true|false>
    }
    \end{verbatim}
    \end{minipage} \\
    \bottomrule
    \end{tabular}
\end{table}
\clearpage
\twocolumn

\clearpage
\onecolumn
\begin{table}[h]
    \centering
    \caption{Stage 3 (Refinement) Prompt}
    \label{tab:prompt_stage3}
    \begin{tabular}{p{0.95\textwidth}}
    \toprule
    \textbf{Prompt:} \\
    \midrule
    You refine the bounding box of the main sound-emitting object by integrating IMAGE, AUDIO, and Stage 2 analysis results. \\
    \\
    \textbf{Context:} \\
    - previous\_bbox \\
    - audio\_class \\
    - image size \(W \times H\) \\
    - av\_consistency, role\_tags, anchor\_votes, keep \\
    \\
    \textbf{Refinement Rules:} \\
    1) Produce a final bbox that best matches the audio class and verified visual anchors, while minimizing unnecessary change. \\
    2) The bbox must remain inside the image bounds \([0, W-1] \times [0, H-1]\) and satisfy \(x1 < x2, y1 < y2\). \\
    3) Unless the previous box is clearly incorrect, limit coordinate adjustments to within ±MAX\_DELTA\_PX per side. \\
    4) Optionally describe the modification using an ``ops" field: delta, expand, shrink, or recenter. \\
    5) Provide a factual refined\_description consisting of 2–4 sentences describing the scene and its relation to the audio class. \\
    \\
    \textbf{STRICT OUTPUT:} \\
    \begin{minipage}{\linewidth}
    \begin{verbatim}
    {
      "bbox": [x1, y1, x2, y2],
      "changed": true/false,
      "ops": {...} | null,
      "refined_description": "..."
    }
    \end{verbatim}
    \end{minipage} \\
    \bottomrule
    \end{tabular}
\end{table}
\clearpage
\twocolumn

\end{document}